\documentclass[letterpaper, 10 pt, conference]{ieeeconf}  

\IEEEoverridecommandlockouts                              
\usepackage{graphicx}\graphicspath{{figure/}}%
\usepackage{epstopdf}%
\usepackage{overpic}%
\usepackage[list=off]{caption}%
\usepackage{fmtcount}%
\usepackage{filecontents}%
\usepackage{xspace}
\usepackage[acronym]{glossaries}%
\glsdisablehyper 
\makeglossaries%

\newglossaryentry{rbf} 
{
	name={RBF},
	description={radial basis function},
	first={\glsentrydesc{rbf} (\glsentrytext{rbf})},
	plural={\glsentrytext{rbf}s},
	descriptionplural={\glsentrydesc{rbf}s},
	firstplural={\glsentrydescplural{rbf} (\glsentryplural{rbf})}
} 

\providecommand{\KCL    }{King's College London}

\providecommand{\CORE   }{the Centre for Robotics Research}

\providecommand{\DOE   }{Department of Engineering}

\providecommand{\EPSRC  }{Engineering and Physical Sciences Research Council}

\newacronym{pd}{PD} {Parkinson's Disease}%
\newacronym{mi}{MI} {mutual information}%
\newacronym{svm}{SVM}{support vector machine}%
\newacronym{pcc}{PCC}{Pearson’s correlation coefficient}%
\newacronym{ar}{AR}{activity recognition}%
\newacronym{me}{ME}{motion estimation}%
\newacronym{gmm}{GMM}{Gaussian mixture model}%
\newacronym{em}{EM}{expectation maximization}%
\newacronym{pdf}{PDF}{probability density function}%
\newacronym{js}{JS}{Jensen-Shannon divergence}%
\newacronym{kl}{KLD}{Kullback-Leibler divergence}%
\newacronym{cdf}{CDF}{cumulative distribution function}%
\newacronym{ks}{KS}{Kolmogorov-Smirnov}%
\newacronym{mse}{MSE}{mean squared error}%
\newacronym{etextile}{e-textile}{electronic textile}%
\newacronym{rmse}{RMSE}{root mean squared error}%
\newacronym{nsl}{NSL}{neural structured learning}%
\newacronym{knn}{KNN}{K-nearest neighbors}%
\newacronym{dt}{DT}{decision tree}%
\newacronym{cnn}{CNNs}{convolutional neural networks}%
\newacronym{nn}{NN}{neural networks}%
\newacronym{dft}{DFT}{discrete fourier transform}%
\newacronym{drm}{DRM}{discriminative regression machines}%
\newacronym{kde}{KDE}{kernel density estimation}%
\newacronym{hmm}{HMM}{Hidden Markov model}%
\newacronym{lr-hmm}{LR-HMM}{Left-to-Right Hidden Markov model}%
\newacronym{lr-hmms}{LR-HMMs}{Left-to-Right Hidden Markov models}%

\newacronym{gmr}{GMR}{Gaussian mixture regression}%
\newacronym{htf}{HTF}{human trajectory forecasting}%
\newacronym{lr-cdhmm}{LR-CDHMM}{left-to-right continuous density hidden Markov model}%
\newacronym{dtw}{DTW}{dynamic time warping}%
\newacronym{stomp}{STOMP}{stochastic trajectory optimization for motion planning}%
\newacronym{ptp}{PTP}{point-to-point}%
\newacronym{gp}{GP}{Gaussian process}%
\newacronym{emg}{EMG}{electromyogram}%
\newacronym{imu}{IMU}{Inertial Measurement Unit}%
\newacronym{hrc}{HRC}{Human-Robot Collaboration}%
\glsunset{etextile}%
\usepackage{amssymb}
\usepackage{mathtools}
\usepackage{amsmath}
\usepackage{gensymb}
\usepackage{amsfonts}

\mathchardef\mhyphen="2D   

\providecommand{\R}     {\mathbb{R}}          
\providecommand{\N}     {\mathcal{N}}         

\providecommand{\estimated} [1]{\tilde{#1}}

\providecommand{\nd}      {n}                              
\providecommand{\Nd}      {\mathcal{\MakeUppercase{\nd}}}  
\providecommand{\nk}      {k}                              
\providecommand{\Nk}      {\mathcal{\MakeUppercase{\nk}}}  

\providecommand{\x}      {x}                  

\providecommand{\btheta}  {\boldsymbol{\theta}}               

\providecommand{\bA}     {\mathbf{A}}         
\providecommand{\bb}     {\mathbf{b}}         
\providecommand{\bB}     {\mathbf{B}}         

\providecommand{\u}      {u}                  

\providecommand{\nu}     {q}                  

\renewcommand  {\c}         {c}               
\providecommand{\Prob}	    {P} 			

\providecommand {\tm}        {m} 
\providecommand{\Tm}     {\mathcal{\MakeUppercase{\tm}}} %
\providecommand{\tv}     {v} %
\providecommand{\Tv}     {\mathcal{\MakeUppercase{\tv}}} %
\providecommand{\tq}     {q} %
\providecommand{\Tq}     {\mathcal{\MakeUppercase{\tq}}} %

\providecommand {\df}        {F}
\providecommand {\dr}        {R}
\providecommand {\fk}        {f_s} 
\providecommand{\tk}{t}    
\providecommand {\Tk}        {T} 
\providecommand{\TK}      {\mathcal{\MakeUppercase{\Tk}}}
\providecommand {\To}        {O} 
\providecommand {\xo}        {o} 
\providecommand {\ty}        {Y} 
\providecommand {\Ts}        {S} 
\providecommand {\ts}        {s} 
\providecommand {\tx}        {x} 
\providecommand {\ttx}        {\mathbf{\mathbf{y}}} 

\providecommand{\B}     {B} 
\providecommand{\bb}     {b} 
\providecommand{\ba}     {a} 
\providecommand{\nn}     {n} %
\providecommand{\Nn}     {\mathcal{\MakeUppercase{\nn}}} %
\providecommand{\Tpi}     {\mathcal{\pi}} %

\renewcommand{\xo}{\mathbf{\ty}}%
\renewcommand{\To}{\mathcal{\mathbf{\MakeUppercase{\tx}}}}%

\providecommand{\elambda}{\estimated{\lambda}}%

\FPset {\NdscotchYokeExperiment}{50}%
\usepackage{xspace}
\usepackage{siunitx} 
\usepackage{nameref}

\providecommand{\eg}{\textit{e.g.,}~} %
\providecommand{\ie}{\textit{i.e.,}~} %
\providecommand*{\sref}[1]{\S\ref{s:#1}}            
\providecommand{\tablename}{Table}
\providecommand{\figurename}{Fig.}
\providecommand*{\fref}[1]{\figurename~\ref{f:#1}}  
\providecommand*{\eref}[1]{(\ref{e:#1})}            

\let\labelindent\relax 
\usepackage[inline]{enumitem} 
\setlist{nolistsep}
\providecommand{\il}[1]{\begin{enumerate*}[label=(\roman*)]#1\end{enumerate*}} 
\providecommand{\cl}[1]{\begin{enumerate*}[label=(\alph*)]#1\end{enumerate*}}  

\usepackage{xcolor}
\colorlet{xr}{blue}
\colorlet{xf}{red}

\usepackage[normalem]{ulem}                                        
\usepackage{marginnote}%
\usepackage[textwidth=\marginparwidth,colorinlistoftodos]{todonotes}%
\providecommand{\tinytodo}[2][]%
{\todo[caption={#2}, size=\small, #1]{\renewcommand{\baselinestretch}{0.6}\selectfont#2\par}}%
\colorlet{jb}{red}%
\colorlet{MH}{red}%
\providecommand{\commentcolourcode}{Comments colour code: %
{\color{MH}MH}} %
\providecommand  {\colorsout}[1]{\bgroup\markoverwith{\textcolor{#1}{\rule[0.5ex]{2pt}{0.4pt}}}\ULon} 
\providecommand{\atJB} {{\color{jb}@JB}}%
\providecommand  {\done}   [1]{\sout{#1}}%
\providecommand  {\edit}   [3]{\colorsout{#3}{#1}{\color{#3}{#2}}}%
\providecommand  {\nomment}[2]{\tinytodo[color=white,nolist,linecolor=#2,bordercolor=white,noinline]{\protect{\scriptsize\color{#2}#1}}}%
\providecommand  {\note}[2]{\tinytodo[color=white,linecolor=#2,bordercolor=white,noinline]{\protect{\scriptsize\color{#2}#1}}}%
\providecommand{\makecomments}{%
\section*{Usage notes}~\\
\noindent Use \texttt{\textbackslash note\{your comment\}\{your initials\}} to add comments/to dos. For example,
\underline{J}oe \underline{B}loggs adds comments using \texttt{\textbackslash note\{Comment.\}\{jb\}}.\nomment{These comments will appear as margin notes. They will also appear in the todo list on the first page.}{jb} \\[2ex]
You can mark a comment as resolved using \texttt{\textbackslash done\{Comment text.\}}, (\eg \texttt{\textbackslash note\{\textbackslash done\{\textbackslash atJB: Please fix this.\}\}\{MH\}}). It will then be formatted like this: \done{\atJB: Please fix this.}.\\[2ex]
\commentcolourcode\\[1ex]
\printacronyms~\\[2ex]
\listoftodos~\\[2ex]%
\clearpage\setcounter{page}{1}%
}%
\providecommand{\makenotes}{\clearpage\appendix\input{notes}}%
\providecommand{\makefinal}{%
\renewcommand{\makecomments}{}\renewcommand{\makenotes}{}\renewcommand{\note}[1]{}\renewcommand{\done}[1]{}\renewcommand{\edit}[3]{\#2}%
}%
\usepackage{soul}%
\colorlet{ID}{blue}%
\colorlet{TS}{green}%
\renewcommand{\commentcolourcode}{Comments colour code: %
 {\color{MH}MH}%
,{\color{ID}ID}%
,{\color{TS}TS}%
} 
\providecommand{\ts}[1]{\note{#1}{TS}}%
\makeatletter\newcommand{\manuallabel}[2]{\def\@currentlabel{#2}\label{#1}}\makeatother
\usepackage[%
backend=biber,%
style=ieee,%
url=false,%
doi=false,%
isbn=false,%
doi=true,%
eprint=false,%
]{biblatex}%
\usepackage[top=19.1mm,bottom=19.1mm,left=19.1mm,right=19.1mm]{geometry}%
\usepackage{graphics}%
\usepackage{tikz,pgfplots}%
\usepackage{subfigure}%
\usepgfplotslibrary{statistics}%
\usepackage{booktabs} %
\usepackage{overpic}
\usepackage{siunitx}%
\usepackage{glossaries}
\usepackage{caption}%
\usepgfplotslibrary{groupplots}
\usetikzlibrary {arrows.meta} %
\usepackage{svg}

\xdefinecolor{mycolor}{RGB}{62,96,111} 
\colorlet{bancolor}{mycolor}

\tikzset{
    error band/.style={fill=orange},
    error band style/.style={
        error band/.append style=#1
    }
}
\usepackage{pgfplots}%
\usepgfplotslibrary{fillbetween}
\usetikzlibrary{spy}%
\title{\LARGE \bf%
Human Movement Forecasting with Loose Clothing}%
\author{Tianchen Shen$^{1}$, Irene Di Giulio$^{2}$ and Matthew Howard$^{1}$
\thanks{This work was supported by King's College London, the China Scholarship Council and \EPSRC\ (EP/M507222/1). For the purpose of open access, the authors have applied a Creative Commons Attribution (CC BY) license to any Accepted Manuscript version arising.}
\thanks{$^{1}$Tianchen Shen ({\tt\small tianchen.shen@kcl.ac.uk}) and Matthew Howard are with \CORE, \DOE, \KCL, UK.}%
\thanks{$^{2}$Irene Di Giulio is with Centre for Human and Applied Physiological Sciences, \KCL, UK.}%
}%
\begin{document}%
\makefinal
\makecomments%
\maketitle%
\thispagestyle{empty}%
\pagestyle{empty}%
\begin{abstract}%
Human motion prediction and trajectory forecasting are essential in human motion analysis. Nowadays, sensors can be seamlessly integrated into clothing using cutting-edge electronic textile (e-textile) technology, allowing long-term recording of human movements outside the laboratory. Motivated by the recent findings that clothing-attached sensors can achieve \emph{higher} activity recognition accuracy than body-attached sensors. This work investigates the performance of human motion prediction using clothing-attached sensors compared with body-attached sensors. It reports experiments in which statistical models learnt from the movement of loose clothing are used to predict motion patterns of the body of robotically simulated and real human behaviours. Counter-intuitively, the results show that fabric-attached sensors can have \emph{better} motion prediction performance than rigid-attached sensors. Specifically, The fabric-attached sensor can \emph{improve} the accuracy up to $40\%$ and requires up to $80\%$ \emph{less} duration of the past trajectory to achieve high prediction accuracy (\ie $95\%$) compared to the rigid-attached sensor.%
\end{abstract}%
\section{Introduction}\label{s:introduction}%
Human motion prediction and trajectory forecasting are crucial for understanding human motion in various research areas \cite{kothari2021human}, ranging from human-robot interaction (\eg service robots \cite{rudenko2020human}), human-robot collaboration in manufacturing \cite{liu2019deep} to rehabilitation devices (\eg exoskeleton robots \cite{qiu2020exoskeleton}). With the latest \gls{etextile} technology, sensors can be embedded in clothing \cite{castano2014smart}, ensuring comfortable and unobtrusive wear for users. This allows for the recording of human movement outside the laboratory for long periods \cite{yang2019textiles}. 

However, a challenge in recording human movement using \gls{etextile} is the potential inclusion of motion artefacts caused by the movement of clothing with respect to the body. Many different approaches have been used to address this issue. For example, \il{\item sensors have been tightly attached to the body using tape \cite{qiu2022sensor,banos2014dealing}, \item attached to tight-fitting clothing \cite{li2009novel} or \item statistical machine learning/signal processing methods have been employed to reduce artefacts \cite{michael2014eliminating,michael2015learning,lorenz2022towards}}. However, an increasing body of work \cite{michael2017,jayasinghe2019comparing,jayasinghe2022comparing,jayasinghe2023classification,shen2022probabilistic} suggests fabric motion could in fact be \emph{useful} for human motion recognition. Inspired by this, this paper focuses on \emph{forecasting human motion patterns from loose clothing}.

\fref{simulation} illustrates the concert of the approach taken in this paper. Future movement (dashed line and shaded area) is forecasted based on the motion prediction made using the movement of the clothing during the past period collected from the clothing-attached sensor (solid line) \footnote{In this paper, the term ``motion prediction'' is used to refer to the classification of past movements (\eg distinguishing between walking and running), while ``motion forecasting'' refers to predicting the future time course of a movement \eg the positions of a point of the body over time).}.

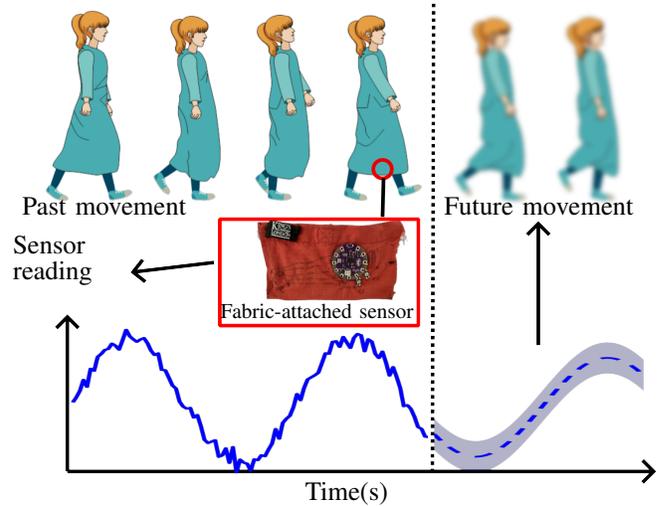
\begin{figure}
\begin{overpic}[width=\linewidth]{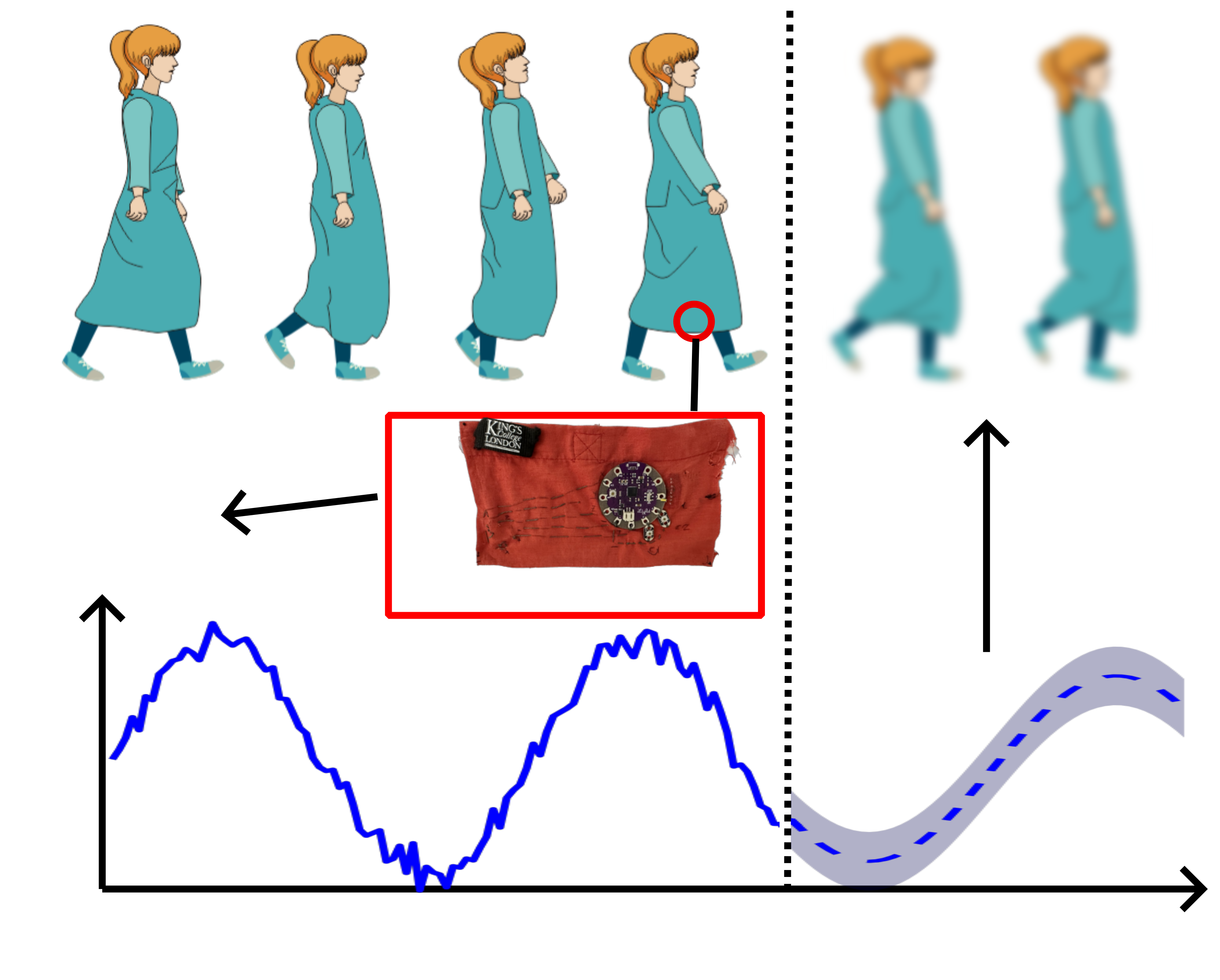}
\put(45,3){Time(s)}
\put(0,41){Sensor}
\put(0,37){reading}
\put(0,47){ Past movement}
\put(65,47){ Future movement}
\put(32,31){\footnotesize Fabric-attached sensor}
\end{overpic}
\caption{Illustration of human movement forecasting. The future human movement (dashed line and shaded area) is forecasted based on the past movement of the clothing-attached sensor (solid line).} 
\label{f:simulation}%
\end{figure}

This paper solves human motion prediction and trajectory forecasting problems using \gls{lr-hmms}. The proposed approach works by eliciting the movement class from the \gls{lr-hmms}, then using those same \gls{lr-hmms} to forecast the movement. The experimental results suggest models trained on data from clothing-attached sensors can lead to \emph{higher} prediction accuracy and require \emph{less} time to achieve high accuracy compared to those trained on data from rigidly-attached sensors. Once the correct motion is predicted, the future trajectory can be forecasted using the past trajectory recorded by a rigid-attached sensor. Furthermore, the statistical distance (\ie cross-fitness distance) between \gls{lr-hmms} is used to estimate discrimination information and understand this counterintuitive phenomenon.%
%
%
\section{Problem Definition}\label{s:problem_definition}%
\renewcommand{\lambda}{\btheta}%
The approach proposed in this paper aims to forecast future human body movement $\overset{\rightarrow}{\xo^r}=[\ttx_{\tv+1}^r,\ttx_{\tv+2}^r\ldots\ttx_{\Tv}^r]$ based on the movement of clothing $\overset{\leftarrow}{\xo^f} = [\ttx_{1}^f,\ttx_{2}^f\ldots\ttx_{\tv}^f]$ using a statistical model trained on past movements.

The latter is trained on data in which body and clothing-attached sensors are used to simultaneously record movement for $\Tv$ time steps with $\Tm$ features as a person performs $\Nk$ repetitions of a given motion of a predefined class. For example, the sensors may record the multi-dimensional orientations of the human body and clothing when a person is walking. Each movement recorded consists of $\Tv$ sensor readings, \ie $\xo_{\nk} = [\ttx_{1},\ttx_{2}\ldots\ttx_{\Tv}]\in\R^{\Tm\times\Tv}$, and is assumed to belong to one of a finite, discrete set of motions/classes (\eg walking, running)\footnote{Throughout the paper, without loss of generality, all prediction tasks are assumed to be binary.}. The complete set of recordings consist of the body- $\To^r=[\xo_1^r,\xo_2^r,\dots\xo_{\Nk}^r]\in\R^{\Tm\times\Nk\times\Tv}$ and the clothing-attached sensor readings $\To^f=[\xo_1^f,\xo_2^f,\dots\xo_{\Nk}^f]\in\R^{\Tm\times\Nk\times\Tv}$. At deployment (\ie once the model has been trained) it is assumed that \emph{no further direct sensing of the body is available}, \ie motion classification and forecasting rely solely on data from fabric-attached sensors.



%
%
%
\section{Methodology}\label{s:hmm}%
\subsection{\Acrlong{lr-hmm} parameter set up}
An \gls{hmm} is used to predict the motion of the past human body $\overset{\leftarrow}{\xo^r}$ and the clothing movement $\overset{\leftarrow}{\xo^f}$ and formulate the probabilistic trajectory model using human body movement $\To^r$. \gls{hmm}s are suitable for modelling time series data in a wide range of applications \cite{visser2011seven}. This is because they are easy to build and manipulate, and optimal algorithms exist to train and score them, such as the forward algorithm and the Viterbi algorithm \cite{papageorgiou2014hidden}.

The following introduces the set-up of the \gls{hmm}s \cite{rabiner1985some}. 
Consider an $\Nn$-state Markov chain where the individual states are denoted as $\Ts =\{\ts_{1},\ts_{2},\ldots,\ts_{\Nn}\}$. The sequence of states that follows the chronological order of the sensor readings is defined as the most likely state sequence, denoted as $\Tq=\{\tq_{1},\tq_{2},\ldots, \tq_{\Tv}\}$. The most likely state at time step $\tv$ is $\tq_{\tv}$. The estimated value in the future trajectory is in the continuous measurement space.

The transitions between these individual states are defined by the transition probability matrix $\bA=\{\ba_{ij}\}$, whose elements are 
\begin{equation}
    \ba_{ij}= \Prob(\tq_{\tv}=\ts_{j}|\tq_{\tv-1} =\ts_{i}), \quad 1\leq i,j\leq\Nn.
\end{equation}
$\ba_{ij}$ represents the probability of transitioning to state $\ts_{j}$ at time step $\tv$, conditional on being in state $\ts_{i}$ at time step $\tv-1$\footnote{As the sensor reading progresses with each time step, the probability of transitioning to state $\ts_{j}$ at time $\tv$, given that it was in state $\ts_{i}$ at time step $\tv-1$, is much higher than remaining in state $\ts_{i}$ at time step $\tv$. (\ie $ \ba_{ij}>> \ba_{ii}$).}.

The emission probability, denoted as $\bB$, is defined as follows: $\bb_{i}(\xo_{\tv})$ represents the probability of an individual state $\ts_{i}$ given sensor readings $\xo_{\tv}$ at time step $\tv$. It is defined by the Gaussian probability density function:
\begin{equation}\label{e:Bt}%
\bb_{i}(\xo_{\tv})  =\Prob(\xo_{\tv}|\tq_{\tv} = \ts_{i}) =\mathcal \N(\mu_{i},\Sigma_{i}),1\leq i\leq\Nn.
\end{equation}%

The initial (\ie $\tv=1$) state distribution is 
\begin{equation}\label{e:pii}
	\pi_{i} =\Prob(\tq_{1}=\ts_{i}),\qquad1\leq i\leq\Nn.
\end{equation}
%

The number of individual states is set to be equal to the number of time steps in each observation trajectory (\ie $\Nn =\Tv$). This implies that each state represents a specific time step in observation. This choice is designed to maximise the performance of the forecasting. This type of \gls{hmm} is also known as a \gls{lr-hmm} \cite{subakan2015method}. A compact notation of the \gls{lr-hmm} parameters is $\lambda = \{\Tpi,\bA ,\bB\}$. 

    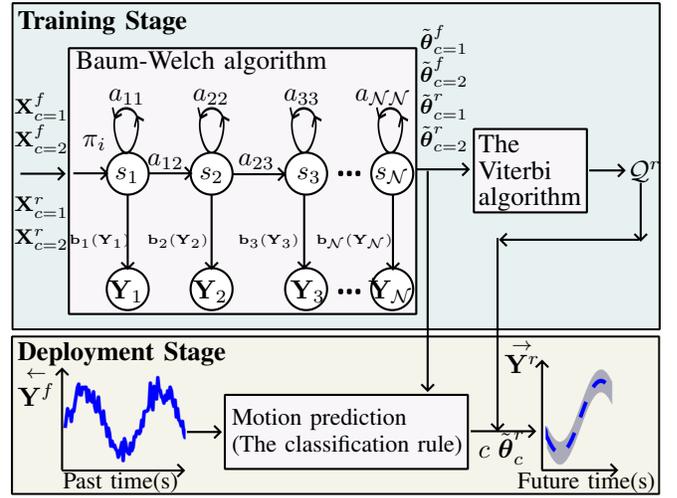
\begin{figure}
    \centering
\begin{overpic}[width=1\linewidth]{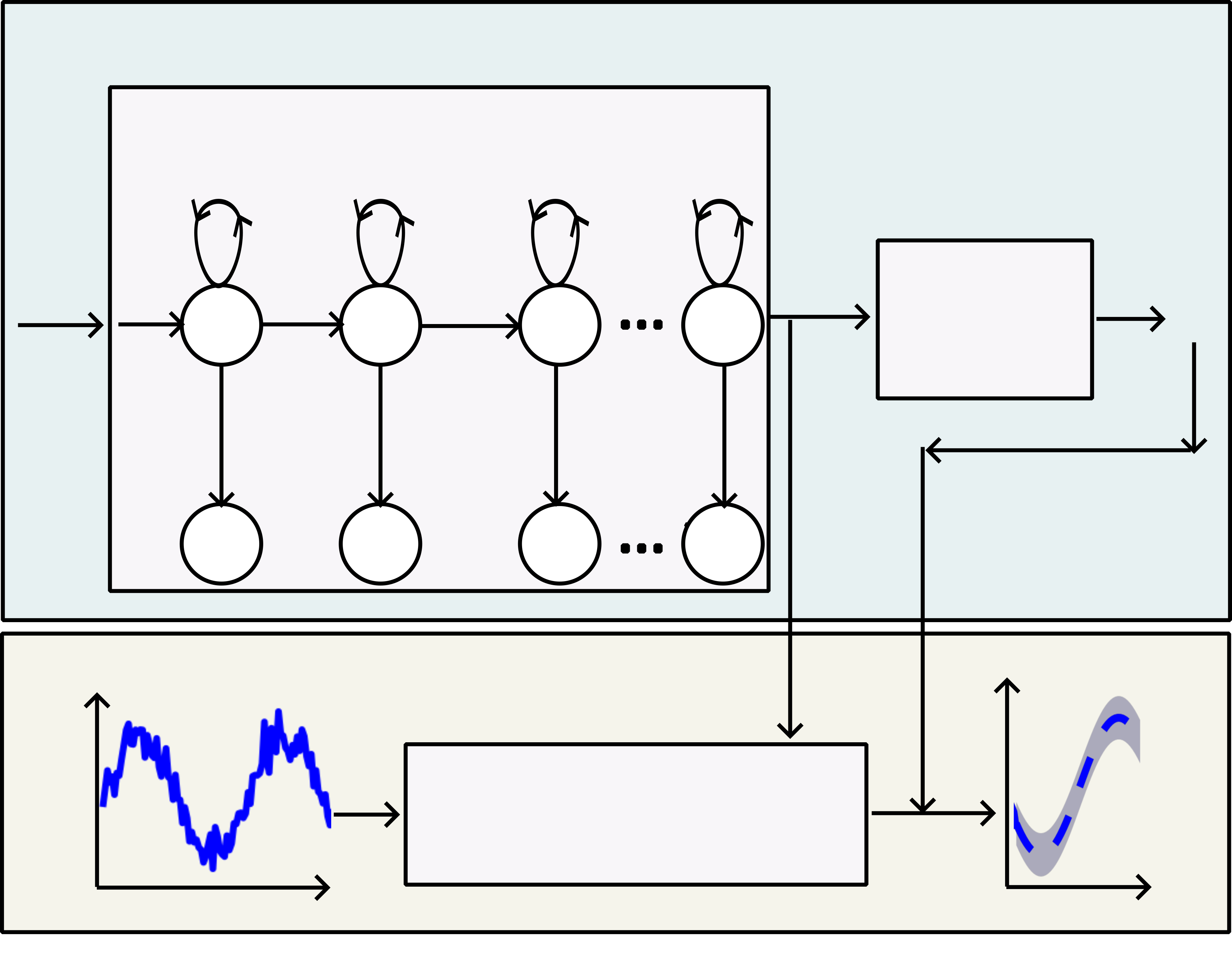}
\begin{tikzpicture}
\coordinate (ref) at (0,0);
\put (0.5,46){\footnotesize $\To_{\c=1}^r$}
\put (0.5,41){\footnotesize $\To_{\c=2}^r$}
\put (0.5,56){\footnotesize $\To_{\c=2}^f$}
\put (0.5,61){\footnotesize $\To_{\c=1}^f$}
\put (11,56){ $\pi_{i}$}
\put (15,63){ $\ba_{11}$}
\put (28,63){ $\ba_{22}$}
\put (42,63){ $\ba_{33}$}
\put (53,63){ $\ba_{\Nn \Nn}$}
\put (16,51){ $\ts_{1}$}
\put (29,51){ $\ts_{2}$}
\put (44,51){ $\ts_{3}$}
\put (56,51){ $\ts_{\Nn}$}
\put (21,53){ $\ba_{12}$}
\put (35,53){ $\ba_{23}$}
\put (15,32.5){ $\xo_{1}$}
\put (28,32.5){ $\xo_{2}$}
\put (43,32.5){ $\xo_{3}$}
\put (55,32.5){ $\xo_{\Nn}$}
\put (9,41){\tiny $\bb_{1}(\xo_{1})$}
\put (21,41){\tiny $\bb_{2}(\xo_{2})$}
\put (35,41){\tiny $\bb_{3}(\xo_{3})$}
\put (47,41){\tiny $\bb_{\Nn}(\xo_{\Nn})$}

\put (63,71){\footnotesize $\elambda_{\c=1}^f$}
\put (63,66){\footnotesize $\elambda_{\c=2}^f$}
\put (63,56){\footnotesize $\elambda_{\c=2}^r$}
\put (63,61){\footnotesize $\elambda_{\c=1}^r$}
\put (95,51){ $\Tq^r$}
\put (72,8){ $\c$}
\put (75,8){ $\elambda_{\c}^r$}

\put (1,16){$\overset{\leftarrow}{\xo^f}$}
\put (76,21){$\overset{\rightarrow}{\xo^r}$}

\end{tikzpicture}
\put(8,3.2){\small Past time(s)}
\put(34,13){\small Motion prediction}
\put(34,9){\small (The classification rule)}
\put(10,68){Baum-Welch algorithm}
\put(1,74){\textbf{Training Stage}}
\put(1,23){\textbf{Deployment Stage}}
\put(72,55){The}
\put(72,51){Viterbi}
\put(72,47){algorithm}

\put(78,3.2){\small Future time(s)}
\end{overpic}
\caption{The framework for motion prediction and trajectory forecasting based on \gls{lr-hmm}. The future human movement $\overset{\rightarrow}{\xo^r}$ (dashed line and shaded area) is forecasted based on the prediction label $\c$ of the past clothing movement $\overset{\leftarrow}{\xo^f}$ (solid line) and the probability trajectory model $\elambda_{c}^r$ formulated by body movement using the Baum-Welch algorithm and the Viterbi algorithm.} 
\label{f:diagram}%
\end{figure}
\subsection{\Acrlong{lr-hmm} parameter estimation}\label{s:hmmparameter}
The Baum-Welch algorithm is used to estimate the \gls{lr-hmm} parameters $\elambda_{\c=1}^r$, $\elambda_{\c=2}^r$, $\elambda_{\c=1}^f$ and $\elambda_{\c=2}^f$, given sensor readings $\To_{\c=1}^r$, $\To_{\c=2}^r$, $\To_{\c=1}^f$, $\To_{\c=2}^f$.

This method is used to adjust the \gls{lr-hmm} to maximise $\Prob(\To|\lambda)$ \ie
\begin{equation}
	\elambda = \underset{\lambda}{\operatorname{argmax}}\sum_{\nk=1}^{\Nk}\log \Prob (\xo_{\nk}|\lambda).
\label{e:distance}
\end{equation}
See \cite{rabiner1989tutorial} for details.
%
%
%

%

\subsection{Motion prediction}\label{s:class}
The classification rule introduced in \cite{bicego2004similarity} is used to make a prediction using the rigid-attached sensor and the fabric-attached sensor. This compares likelihoods based on two \gls{lr-hmm} parameters and the past trajectory. \ie
\begin{equation}\label{e:c}
\begin{split}
& \c = \underset{\c}{\operatorname{argmax}} \log \Prob(\overset{\leftarrow}{\xo^f}|\elambda_{\c}^f), \quad \c\in\{1,2\}\\
& \c = \underset{\c}{\operatorname{argmax}} \log \Prob(\overset{\leftarrow}{\xo^r}|\elambda_{\c}^r), \quad \c\in\{1,2\}
 \end{split}
\end{equation}
The forward algorithm, as described in \cite{rabiner1989tutorial}, is used to compute the likelihoods of the past trajectory given \gls{lr-hmm} parameters.
\subsection{Trajectory forecasting}\label{s:hmmmost}
%
%

In this paper, each individual state is set to correspond to a specific time step in the observation trajectory. However, it is not explicitly determined which individual state in the estimated \gls{lr-hmm} corresponds to which time step in the observation trajectory. Formulating the probabilistic trajectory model requires finding which individual state corresponds to each time step of the observation trajectory.
%
Therefore, the most likely state sequence $\Tq^r$  needs to be solved using the Viterbi algorithm, as described in \cite{rabiner1989tutorial}. After that, the probability density function of each individual state given the rigid-attached sensor's reading $\xo_{\tv}^r$ at each time step $\tv$ (\ie emission probability $\bB$) and the most likely state sequence $\Tq^r$ can be used to form the probabilistic trajectory model (\ie $\bb_{\tq_{1}}(\xo_{1}^r),\bb_{\tq_{2}}(\xo_{2}^r),\bb_{\tq_{3}}(\xo_{3}^r)\ldots\bb_{\tq_{\Tv}}(\xo_{\Tv}^r)$).

To conclude, in the training stage, given the observations (\ie rigid and fabric-attached sensor readings) of two classes of movements $\To_{\c = 1}^f$, $\To_{\c = 2}^f$, $\To_{\c = 1}^r$ and $\To_{\c = 2}^r$, the \gls{lr-hmm}s parameters $\elambda_{\c=1}^f$, $\elambda_{\c=2}^f$, $\elambda_{\c=1}^r$ and $\elambda_{\c=2}^r$ are estimated using the Baum–Welch algorithm, the most likely state sequence $\Tq^r$ is estimated using the Viterbi algorithm to formulate the probabilistic trajectory model of the rigid movement.

In the deployment stage, the past fabric movement $\overset{\leftarrow}{\xo^f}$ (solid line, see \fref{diagram}) is then used to recognise the movement (\ie $\c = 1$ or $\c = 2$ is determined) using the classification rule, and based on this, the probabilistic trajectory model of the rigid movement is used to forecast the future rigid movement $\overset{\rightarrow}{\xo^r}$ (dashed line and shaded area, see \fref{diagram}).

\section{Experiments}\label{s:experiment}%

\newcounter{casestudy}
\stepcounter{casestudy}%
\subsection{Case Study \thecasestudy: Linear and curved \acrlong{ptp} movements}\label{s:kuka}
\begin{figure}
\begin{overpic}[width=\linewidth]{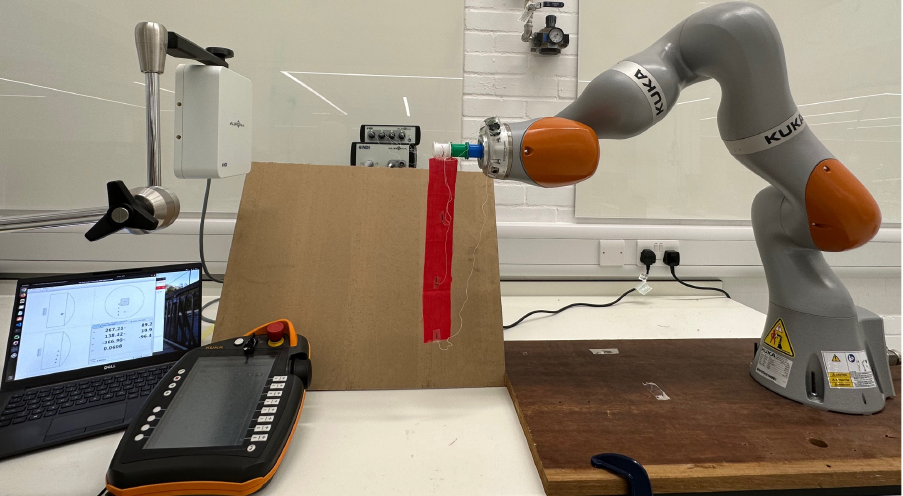}
\begin{tikzpicture}
\coordinate (ref) at (0,0);
\draw [color=green,very thick] (ref)+(1.6,3) rectangle (2.3,4);
\draw [color=green,very thick] (ref)+(0,1.2) rectangle (1.5,2.2);
\draw [color=green,very thick] (ref)+(1,0) rectangle (2.8,1.2);
\draw [color=green,very thick] (ref)+(3.2,3) rectangle (3.9,3.6);
\draw [color=green,very thick] (ref)+(4.4,0.9) rectangle (8.5,4.6);
\draw[stealth-,line width = 3pt,green] (ref)+(4.4,0.9) -- (4.4,0.2) node[right,black,fill=white]{Kuka LBR iiwa robot arm};
\draw[stealth-,line width = 3pt,green] (ref)+(3.3,3) -- (3,2.6) node[left,black,fill=white]{Aurora components};
\draw[stealth-,line width = 3pt,green] (ref)+(1.8,4) -- (1.8,4.4) node[right,black,fill=white]{Field generator};
\draw[stealth-,line width = 3pt,green] (ref)+(1.5,1.7) -- (1.9,1.7) node[right,black,fill=white]{Laptop};
\draw[stealth-,line width = 3pt,green] (ref)+(2.8,0.6) -- (3.2,0.6) node[right,black,fill=white]{Panel};
\draw[stealth-,line width = 3pt,green] (ref)+(4.55,3.2) -- (5,2.5) node[right,black,fill=white]{End effector};
\node at (3.9,3.7)[right,black,fill=white]{$\dr_{1}$};
\node at (3.25,2.6)[right,black,fill=white]{$\df_{2}$};
\node at (3.25,2.05)[right,black,fill=white]{$\df_{3}$};
\node at (3.25,1.5)[right,black,fill=white]{$\df_{4}$};
\filldraw[blue] (4.1,3.2) circle (2pt) node[anchor=west]{};
\filldraw[red] (4.1,2.6) circle (2pt) node[anchor=west]{};
\filldraw[brown] (4.1,2.1) circle (2pt) node[anchor=west]{};
\filldraw[black] (4.1,1.5) circle (2pt) node[anchor=west]{};
\node at (0.2,0.2)[fill=black,text=white,opacity=.25,text opacity=1]{\ref{f:setup-a}};
\end{tikzpicture}
\end{overpic}
\centering
\input{figure/ptp}
\input{figure/circle}
\begin{tikzpicture}
    \begin{groupplot}[
        group style={
            group size=2 by 1,
            horizontal sep=0.8cm,
        },
        width=0.575\linewidth,
        height=0.5\linewidth,
        xmin=0.025, xmax=1,
        ymin=-30, ymax=45,
        xtick={0.025,1},
        ytick={-30,45},
        xticklabel style={
            anchor=near xticklabel,
            /pgf/number format/precision=3,
            /pgf/number format/fixed },
      y label style={at={(axis description cs:0.3,.5)},anchor=south},
               xlabel style={at={(axis description cs:0.5,0.2)},anchor=north}
    ]
    
    \nextgroupplot[xlabel={Time (s)}, ylabel={Orientation($^\circ$)}]
    \addplot[blue, line width=1.5pt] plot [error bars/.cd, y dir=both, y explicit] table [y] {o1.dat};\label{plots:plot20}
    \addplot[red, line width=1.5pt] plot [error bars/.cd, y dir=both, y explicit] table [y] {o2.dat};\label{plots:plot30}
    \addplot[blue!50] plot [error bars/.cd, y dir=both, y explicit] table [y] {o3.dat};\label{plots:plot3}
    \addplot[red!50] plot [error bars/.cd, y dir=both, y explicit] table [y] {o4.dat};\label{plots:plot4}

    \nextgroupplot[xlabel={Time (s)},ytick=\empty]
    \addplot[blue, line width=1.5pt] plot [error bars/.cd, y dir=both, y explicit] table [y] {oo1.dat};
    \addplot[red, line width=1.5pt] plot [error bars/.cd, y dir=both, y explicit] table [y] {oo2.dat};
    \addplot[blue!50] plot [error bars/.cd, y dir=both, y explicit] table [y] {oo3.dat};
    \addplot[red!50] plot [error bars/.cd, y dir=both, y explicit] table [y] {oo4.dat};
    \end{groupplot}
    
  \path (group c1r1.south west) -- (group c2r1.south east)
    node[midway, below, yshift=-7mm] {
        \begin{tabular}{ll}
          \ref{plots:plot20} $\dr_{1}(\c=1)$ & \ref{plots:plot30} $\dr_{1}(\c=2)$ \\
          \ref{plots:plot3} $\df_{4}(\c=1)$ & \ref{plots:plot4} $\df_{4}(\c=2)$ \\
        \end{tabular}
      };
  
      \node at (-0.3,-0.6){\ref{f:setup-b}};
      \node at (4,-0.6){\ref{f:setup-c}};

\end{tikzpicture}
\begin{tikzpicture}[remember picture,overlay]
    \coordinate (bottomLeft) at ([xshift=-104mm,yshift=-66mm]current page.north east); 
    \coordinate (topRight) at ([xshift=-59mm,yshift=-145mm]current page.north east); 
    \draw[line width=1pt, rounded corners=1pt, draw=black] (bottomLeft) rectangle (topRight);
\end{tikzpicture}
\begin{tikzpicture}[remember picture,overlay]
    \coordinate (bottomLeft) at ([xshift=-20mm,yshift=-66mm]current page.north east); 
    \coordinate (topRight) at ([xshift=-59mm,yshift=-145mm]current page.north east); 
    \draw[line width=1pt, rounded corners=1pt, draw=black] (bottomLeft) rectangle (topRight);
\end{tikzpicture}
\caption{\cl{ \item\label{f:setup-a}Experimental setup. The robot arm with a piece of fabric attached to the end effector. The actual moving trajectories and the longitudinal/X-axis (roll) orientation of the rigid-attached sensor and the fabric-attached sensor during \item\label{f:setup-b} linear and \item\label{f:setup-c} curved \gls{ptp} movement. Note that the variations of rigid movements in both types of movements are less obvious.}}
\label{f:setup}%
\end{figure}%

The aim of the first evaluation is to estimate the performance of the proposed approach in recognising and forecasting simple movement patterns. For this, in the first case study, a robotic simulation of human movement is used, in order to eliminate variability in the motion pattern that may confound the experimental results (real human movement is addressed in \sref{human} below). To this end, a robot arm with a piece of fabric attached is used as a proxy for the human arm, to collect reliable and repeatable movements.
\subsubsection{Materials and Methods}
KUKA's LBR iiwa robot arm (KUKA, Germany) is used to execute \gls{ptp} movement encompassing both linear (\texttt{SLIN}) and curved (\texttt{SCIRC}) motion, using KUKA Sunrise.OS $1.11$. \fref{setup}\ref{f:setup-a} shows the experimental set up, comprising a \SI{30}{\cm}$\times$\SI{5}{\cm} strip of woven cotton fabric attached at the end effector of the robot arm. Three sensors (NDI Aurora Magnetic Tracking device, NDI, Canada) are mounted on this fabric strip to record positions and orientations synchronously (both are in three dimensions) at $\fk=\SI{40}{\Hz}$. The sensors are attached along the length of the fabric \il{\item $\SI{20}{\cm}$ ($\df_{2}$), \item $\SI{30}{\cm}$ ($\df_{3}$) and \item $\SI{40}{\cm}$ from the fulcrum ($\df_{4}$) (\ie at the tip of the fabric)}. It is helpful to understand how different sensor placements on the piece of the fabric influence motion prediction accuracy. A sensor ($\dr_{1}$) is rigidly attached to the end effector of the robot arm at the fabric attachment point. The end effector of the robot arm moves at $2.25\si{\cm\per\second}$ in linear and curved \gls{ptp} movements. The trajectory of the rigid-attached sensor ($\dr_{1}$ thick and dark line) and the fabric-attached sensor ($\df_{4}$ thin and light line) during linear and curved \gls{ptp} movements are shown in \fref{setup}\ref{f:setup-b} and \ref{f:setup-c}, respectively. The colours red and blue represent two classes of movement to be predicted. In both types of movement, the variations in rigid movements are less pronounced because the robot arm moves very smoothly.
With this set up, $\Nk=\NdscotchYokeExperiment$ trajectories of length $\Tk=\SI{3.5}{\second}$ at each class of movement are recorded ($\Tv = \fk\times\Tk= 40\times 3.5 = 140$ data points). Motion predictions are made based on the orientations (\ie $\Tm = 3$) of each sensor's readings $\To_{\c=1}^r$, $\To_{\c=2}^r$, $\To_{\c=1}^f$ and $\To_{\c=2}^f$ since the widely used \gls{imu}s can also receive orientation data. $49$ trajectories are randomly chosen from each class of movement of each sensor used to train the separate \gls{lr-hmm}s $\elambda_{\c=1}^r$, $\elambda_{\c=2}^r$, $\elambda_{\c=1}^f$ and $\elambda_{\c=2}^f$. The remaining trajectories are reserved for testing purposes. Motion prediction is decided using equation \eref{c}. The above-described process is repeated $100$ times. The prediction accuracy is calculated as
\begin{equation}
\text{Prediction accuracy} = \frac{\text{number of correct predictions}}{\text{total number of predictions}}.
    \label{e:accuracy}
\end{equation}
The motion prediction accuracy of each sensor is evaluated separately. Once the label of the past movement is predicted, the \gls{lr-hmm} trained using the positions of sensors $\dr_{1}$ and the Viterbi algorithm is used to forecast the future robot movement. To understand the reason why fabric-attached sensors have higher motion prediction accuracy, the information content of the rigid and fabric movement is investigated. The statistical distance between \gls{lr-hmm}s given two categories of movements is computed as a measure of discrimination information between two classes of \gls{hmm}s \cite{topsoe2000some}. The cross-fitness distance of rigid movements $D^r$ and fabric movements $D^f$ is computed as proposed in \cite{porikli2004trajectory}. The formula is shown below: 
\begin{equation}
\begin{split}
        D^r &= \log \Prob(\To_{\c=1}^r|\elambda_{\c=1}^r) +
        \log \Prob(\To_{\c=2}^r|\elambda_{\c=2}^r) \\ & -\log \Prob(\To_{\c=1}^r|\elambda_{\c=2}^r)-\log \Prob(\To_{\c=2}^r|\elambda_{\c=1}^r),\\
        D^f &= \log \Prob(\To_{\c=1}^f|\elambda_{\c=1}^f) +
        \log \Prob(\To_{\c=2}^f|\elambda_{\c=2}^f) \\ & -\log \Prob(\To_{\c=1}^f|\elambda_{\c=2}^f)-\log \Prob(\To_{\c=2}^f|\elambda_{\c=1}^f).
\end{split}
    \label{e:distance}
\end{equation}

\subsubsection{Results}
\begin{figure}
 \begin{tikzpicture}
  \begin{axis}[
        ylabel=  Accuracy ($\%$),
        xmin=0.025, xmax=1.2,
        ymin=0.4, ymax=1,
        xtick={0.025,1.2},
        ytick={0.4,1},
        legend style={legend columns=2},
    y label style={at={(axis description cs:0.3,.5)},anchor=south},
               xlabel style={at={(axis description cs:0.5,0.2)},anchor=north}, 
        legend pos=north west,
         xticklabel style=
        {anchor=near xticklabel,
         /pgf/number format/precision=3,
         /pgf/number format/fixed},
width= 0.53\linewidth,%
height = 0.45\linewidth,%
axis lines = left,  
title style={fill=red!20, text=red, draw=black, very thick, rounded corners},
title={Linear \gls{ptp} Movement}]
 \addplot[blue]plot [
error bars/.cd,
y dir=both, y explicit,
] table [y ] {f1.dat};
 \addplot[red]plot [
error bars/.cd,
y dir=both, y explicit,
] table [y ] {f2.dat};
 \addplot[brown]plot [
error bars/.cd,
y dir=both, y explicit,
] table [y ] {f3.dat};
 \addplot[black]plot [
error bars/.cd,
y dir=both, y explicit,
] table [y ] {f4.dat};
\end{axis}
  \draw[<->, ultra thick, green] (1.9,0.1) -- (1.9,0.8);
\node at (0.5,0.4)[right,text=green,opacity=.25,text opacity=1]{increase};
        \draw [<->, ultra thick, red] (2.2,1.7) --  (2.7,1.7);
      \node at (0.5,1.6)[right,text=red,opacity=.25,text opacity=1]{less time};
\end{tikzpicture}
\usepgfplotslibrary{fillbetween}
\begin{tikzpicture}
 \begin{axis}[
        xmin=0.025, xmax=1.5,
        ymin=0.4, ymax=1,
        xtick={0.025,1.5},
        ytick={0.4,1},
        legend style={legend columns=2},
    every axis y label/.style={at={(current axis.north west)},above=1mm},
               xlabel style={at={(axis description cs:0.5,0.2)},anchor=north}, 
        legend pos=north west,
         xticklabel style=
        {anchor=near xticklabel,
         /pgf/number format/precision=3,
         /pgf/number format/fixed},
width= 0.53\linewidth,%
height = 0.45\linewidth,%
  axis lines = left,title style={fill=red!20, text=red, draw=black, very thick, rounded corners},
title={Curved \gls{ptp} Movement}]]
 \addplot[blue]plot [
error bars/.cd,
y dir=both, y explicit,
] table [y ] {s1.dat};
 \addplot[red]plot [
error bars/.cd,
y dir=both, y explicit,
] table [y ] {s2.dat};
 \addplot[brown]plot [
error bars/.cd,
y dir=both, y explicit,
] table [y ] {s3.dat};
 \addplot[black]plot [
error bars/.cd,
y dir=both, y explicit,
] table [y ] {s4.dat};
\end{axis}
  \draw[<->, ultra thick, green] (0.3,0.6) -- (0.3,2);
       \node at (0.3,1)[right,text=green,opacity=.25,text opacity=1]{$40\%$ increase};
        \draw [<->, ultra thick, red] (0.4,2) --  (2.4,2);
      \node at (0.3,1.5)[right,text=red,opacity=.25,text opacity=1]{$80\%$ less time};
\end{tikzpicture}
\usepgfplotslibrary{fillbetween}
\begin{tikzpicture}
    \begin{axis}[
        xlabel=Time $(s)$,
        ylabel=$D$,
        xmin=0.025, xmax=1.2,
        ymin=-5, ymax=900,
        xtick={0.025,1.2},
        ytick={-5,900},
   y label style={at={(axis description cs:0.3,.5)},anchor=south},
               xlabel style={at={(axis description cs:0.5,0.2)},anchor=north}, 
        legend pos=north west,
         xticklabel style=
        {anchor=near xticklabel,
         /pgf/number format/precision=3,
         /pgf/number format/fixed},
width= 0.525\linewidth,%
height = 0.5\linewidth,%
  axis lines = left]
\addplot [blue]table[x=x,y=y] {df1.dat};
\addplot [red]table[x=x,y=y] {df2.dat};
\addplot [brown]table[x=x,y=y] {df3.dat};
\addplot [black]table[x=x,y=y] {df4.dat};
\addplot [name path=upper,draw=none] table[x=x,y expr=\thisrow{y}+\thisrow{error}/2.5] {df1.dat};
\addplot [name path=lower,draw=none] table[x=x,y expr=\thisrow{y}-\thisrow{error}/2.5] {df1.dat};
\addplot [fill=blue!10] fill between[of=upper and lower];

\addplot [name path=upper,draw=none] table[x=x,y expr=\thisrow{y}+\thisrow{error}/2.5] {df2.dat};
\addplot [name path=lower,draw=none] table[x=x,y expr=\thisrow{y}-\thisrow{error}/2.5] {df2.dat};
\addplot [fill=red!10] fill between[of=upper and lower];

\addplot [name path=upper,draw=none] table[x=x,y expr=\thisrow{y}+\thisrow{error}/2.5] {df3.dat};
\addplot [name path=lower,draw=none] table[x=x,y expr=\thisrow{y}-\thisrow{error}/2.5] {df3.dat};
\addplot [fill=brown!10] fill between[of=upper and lower];

\addplot [name path=upper,draw=none] table[x=x,y expr=\thisrow{y}+\thisrow{error}/2.5] {df4.dat};
\addplot [name path=lower,draw=none] table[x=x,y expr=\thisrow{y}-\thisrow{error}/2.5] {df4.dat};
\addplot [fill=black!10] fill between[of=upper and lower];

\addlegendentry{ $\dr_{1}$};
\addlegendentry{ $\df_{2}$};
\addlegendentry{ $\df_{3}$};
\addlegendentry{ $\df_{4}$};
\end{axis}
  \draw[<->, ultra thick, green] (2.8,0.2) -- (2.8,1.3);
\node at (1.4,1.4)[right,text=green,opacity=.25,text opacity=1]{increase};
\node at (-0.3,-0.6){\ref{f:pred1-a}};%
\end{tikzpicture}
\usepgfplotslibrary{fillbetween}
\begin{tikzpicture}
    \begin{axis}[
        xlabel=Time $(s)$,
        xmin=0.025, xmax=1.2,
        ymin=-5, ymax=700,
        xtick={0.025,1.2},
        ytick={-5,700},
    every axis y label/.style={at={(current axis.north west)},above=1mm},
               xlabel style={at={(axis description cs:0.5,0.2)},anchor=north}, 
        legend pos=north west,
         xticklabel style=
        {anchor=near xticklabel,
         /pgf/number format/precision=3,
         /pgf/number format/fixed},
width= 0.525\linewidth,%
height = 0.5\linewidth,%
  axis lines = left]
\addplot [blue]table[x=x,y=y] {ddf1.dat};
\addplot [red]table[x=x,y=y] {ddf2.dat};
\addplot [brown]table[x=x,y=y] {ddf3.dat};
\addplot [black]table[x=x,y=y] {ddf4.dat};
\addplot [name path=upper,draw=none] table[x=x,y expr=\thisrow{y}+\thisrow{error}/2.5] {ddf1.dat};
\addplot [name path=lower,draw=none] table[x=x,y expr=\thisrow{y}-\thisrow{error}/2.5] {ddf1.dat};
\addplot [fill=blue!10] fill between[of=upper and lower];

\addplot [name path=upper,draw=none] table[x=x,y expr=\thisrow{y}+\thisrow{error}/2.5] {ddf2.dat};
\addplot [name path=lower,draw=none] table[x=x,y expr=\thisrow{y}-\thisrow{error}/2.5] {ddf2.dat};
\addplot [fill=red!10] fill between[of=upper and lower];

\addplot [name path=upper,draw=none] table[x=x,y expr=\thisrow{y}+\thisrow{error}/2.5] {ddf3.dat};
\addplot [name path=lower,draw=none] table[x=x,y expr=\thisrow{y}-\thisrow{error}/2.5] {ddf3.dat};
\addplot [fill=brown!10] fill between[of=upper and lower];

\addplot [name path=upper,draw=none] table[x=x,y expr=\thisrow{y}+\thisrow{error}/2.5] {ddf4.dat};
\addplot [name path=lower,draw=none] table[x=x,y expr=\thisrow{y}-\thisrow{error}/2.5] {ddf4.dat};
\addplot [fill=black!10] fill between[of=upper and lower];
\end{axis}
  \draw[<->, ultra thick, green] (1.5,0.1) -- (1.5,1.3);
 \node at (-0.3,-0.6){\ref{f:pred1-b}};
 \node at (1.6,0.5)[right,text=green,opacity=.25,text opacity=1]{increase};
\end{tikzpicture}
\input{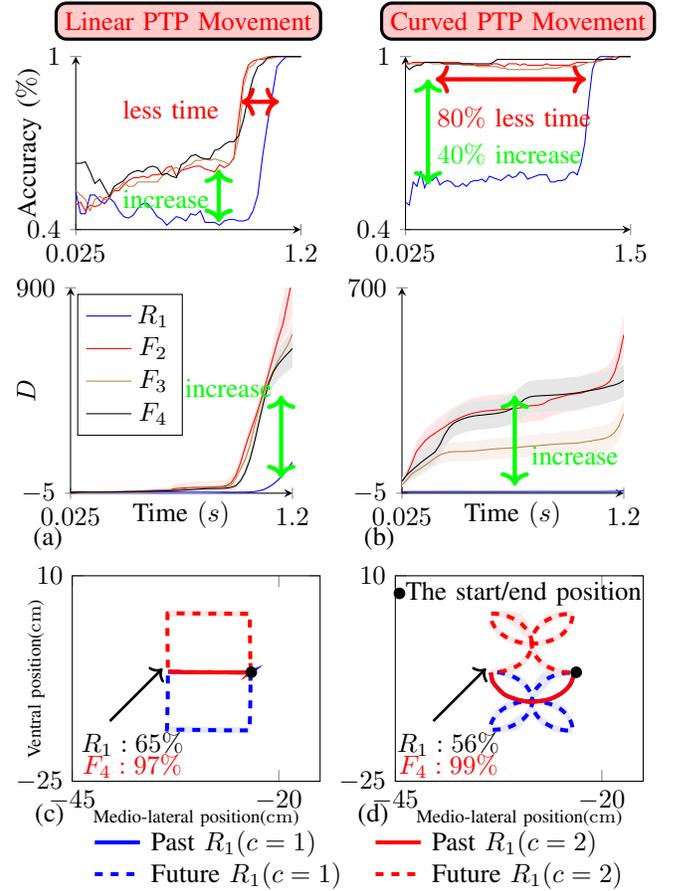}
\caption{The accuracy of motion prediction and the cross-fitness distance from the first time step to $1.2 \si{\second}$ in the contexts of \cl{ \item \label{f:pred1-a} linear and \item \label{f:pred1-b} curved \gls{ptp} movements. Reported are the mean value $\pm$ $\text{s.d}/5$. Panels \item \label{f:pred1-c} and \item \label{f:pred1-d} show the forecasting of linear and curved \gls{ptp} movements, respectively. The future robot movement is forecasted (dashed line) based on $\SI{1}{\second}$ of past movement (solid line). The future trajectory shown is the mean value (dashed line) $\pm$ $\text{s.d}\times5$ (shaded area). } }
    \label{f:pred1}
\begin{tikzpicture}[remember picture,overlay]
    \coordinate (bottomLeft) at ([xshift=-105mm,yshift=-19mm]current page.north east); 
    \coordinate (topRight) at ([xshift=-61mm,yshift=-129mm]current page.north east); 
    \draw[line width=1pt, rounded corners=1pt, draw=black] (bottomLeft) rectangle (topRight);
\end{tikzpicture}
\begin{tikzpicture}[remember picture,overlay]
    \coordinate (bottomLeft) at ([xshift=-19mm,yshift=-19mm]current page.north east); 
    \coordinate (topRight) at ([xshift=-61mm,yshift=-129mm]current page.north east); 
    \draw[line width=1pt, rounded corners=1pt, draw=black] (bottomLeft) rectangle (topRight);
\end{tikzpicture}
\end{figure}

\fref{pred1}\ref{f:pred1-a} and \ref{f:pred1-b} illustrate the motion prediction accuracy and the cross-fitness distance between two classes of movement for linear and curved \gls{ptp} movements, respectively. The accuracy of the clothing-attached sensors is significantly higher than that of the rigidly-attached sensors in both types of movements. The accuracy using fabric movement can increase by around $40\%$ and requires $80\%$ less time length of past movement to get $95\%$ prediction accuracy compared with using rigid movement. The placement of fabric-attached sensors does not appear to have an obvious effect on motion prediction accuracy. The cross-fitness distance of the fabric-attached sensor is higher than the rigid-attached sensor. This indicates that clothing movement provides additional information to simplify the prediction task. \fref{pred1}\ref{f:pred1-c} and \ref{f:pred1-d} show the forecasted movement (dashed line and shaded area) given the past movement (solid line) time step one ($\tv=1$, $\tk=0$) to $\tk=\SI{1}{\second}$ when the robot arm performs linear and curved \gls{ptp} movements, respectively. The motion prediction accuracy using the orientations of rigid and fabric movement at that time is also shown. The motion prediction accuracy of using sensor $\df_{4}$ is higher than using sensor $\dr_{1}$. The dashed line is the future trajectory with the highest probability (\ie the mean of each $\bb_{i}(\xo_{\tv}$). The shaded area represents probability (\ie the standard derivation of each $\bb_{i}(\xo_{\tv}$). Specifically, the further away from the dashed line, the less likely the trajectory is to occur. The thick and light lines are sensor $\dr_{1}$ and $\df_{4}$ movement, respectively.

\stepcounter{casestudy}%
\subsection{Case Study \thecasestudy: Human motion prediction}\label{s:human}
The second evaluation aims to test the performance of the proposed approach on forecasting real human motion \footnote{The experiments reported here were conducted with the ethical approval of King’s College London, UK: MRPP-23/24-40031.}. For this, a human reaching motion is chosen as it is a common daily activity. Specifically, the scenario considered here consists of reaching to press one of a set of buttons, and the task is to predict which button the person intends to hit \emph{before the button is pressed}.
\subsubsection{Materials and Methods}
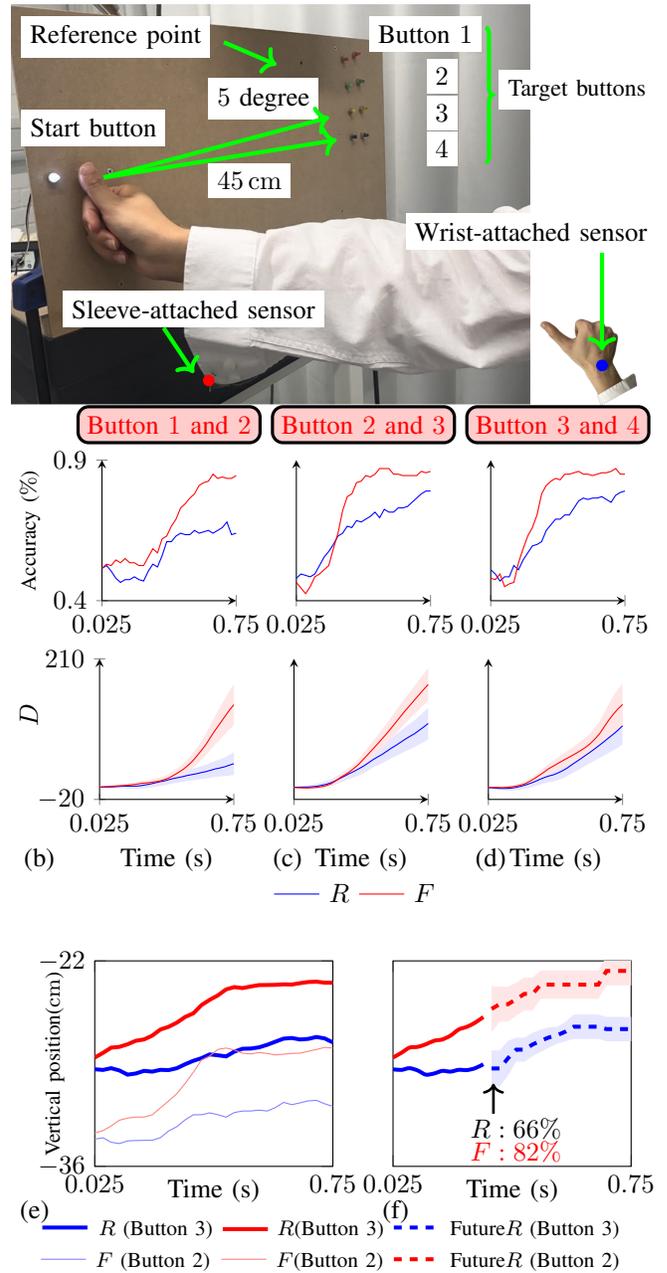
\begin{figure}
\begin{overpic}[width=0.8\linewidth]{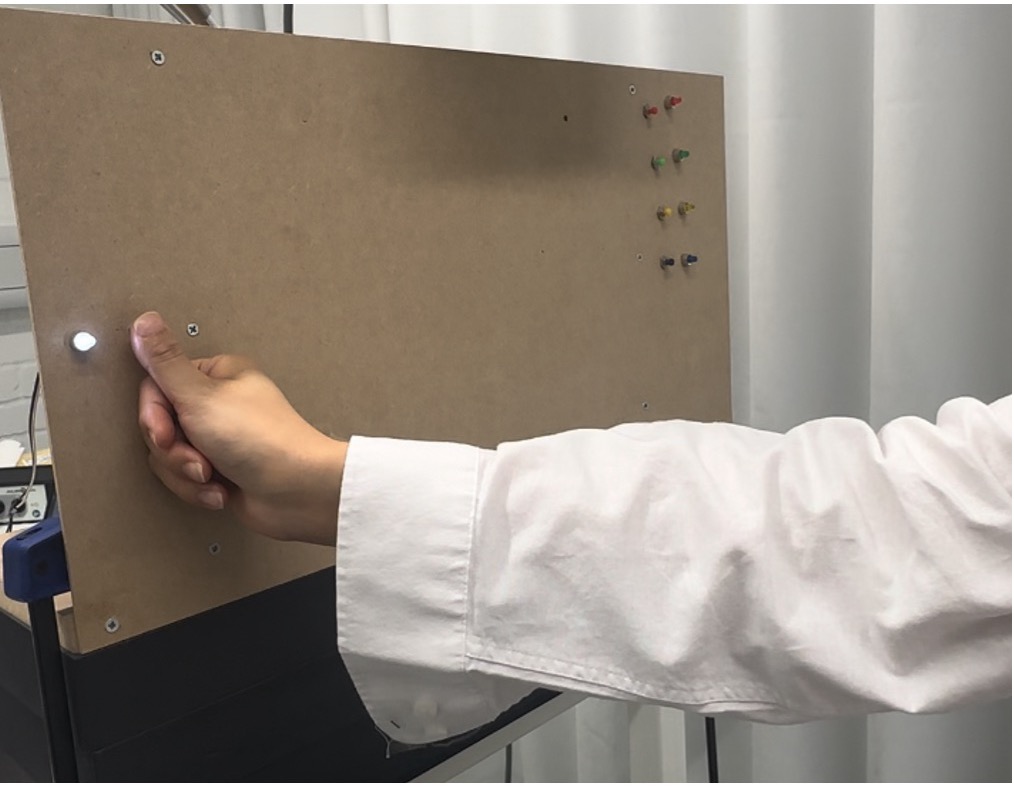}
\put(2,50){\tikz\node[right,black,fill=white] {Start button};}
\put(69,68){\tikz\node[right,black,fill=white] {Button $1$};}
\put(80,60){\tikz\node[right,black,fill=white] { $2$};}
\put(80,53){\tikz\node[right,black,fill=white] { $3$};}
\put(80,46){\tikz\node[right,black,fill=white] { $4$};}
\put(38,40){\tikz\node[right,black,fill=white] { $\SI{45}{\cm}$};}
\put(38,55){\tikz\node[right,black,fill=white] { $5$ degree};}
\put(2,67){\tikz\node[right,black,fill=white] { Reference point};}
\put(10,15){\tikz\node[right,black,fill=white] {Sleeve-attached sensor};}
\put(76,30){\tikz\node[right,black,fill=white] {Wrist-attached sensor};}
\put(94,57){\tikz\node[right,black,fill=white] {\small Target buttons};}
\begin{tikzpicture}
\coordinate (ref) at (0,0);
\draw[->, ultra thick, green] (ref)+(1.2,3) -- (4.3,3.5);
\draw[->, ultra thick, green] (ref)+(1.2,3) -- (4.2,3.8);
\draw[->, ultra thick, green] (ref)+(2,1) -- (2.4,0.4);
\draw[<-, ultra thick, green] (ref)+(3.5,4.5) -- (2.8,4.8);
\draw[decorate,decoration={brace,mirror,raise=8pt},ultra thick,green]  (ref)+(6,3.2) --  (6,5);
\filldraw[red] (2.6,0.3) circle (2pt) node[anchor=west]{};
\end{tikzpicture}
\end{overpic}
\begin{overpic}[width=0.15\linewidth]{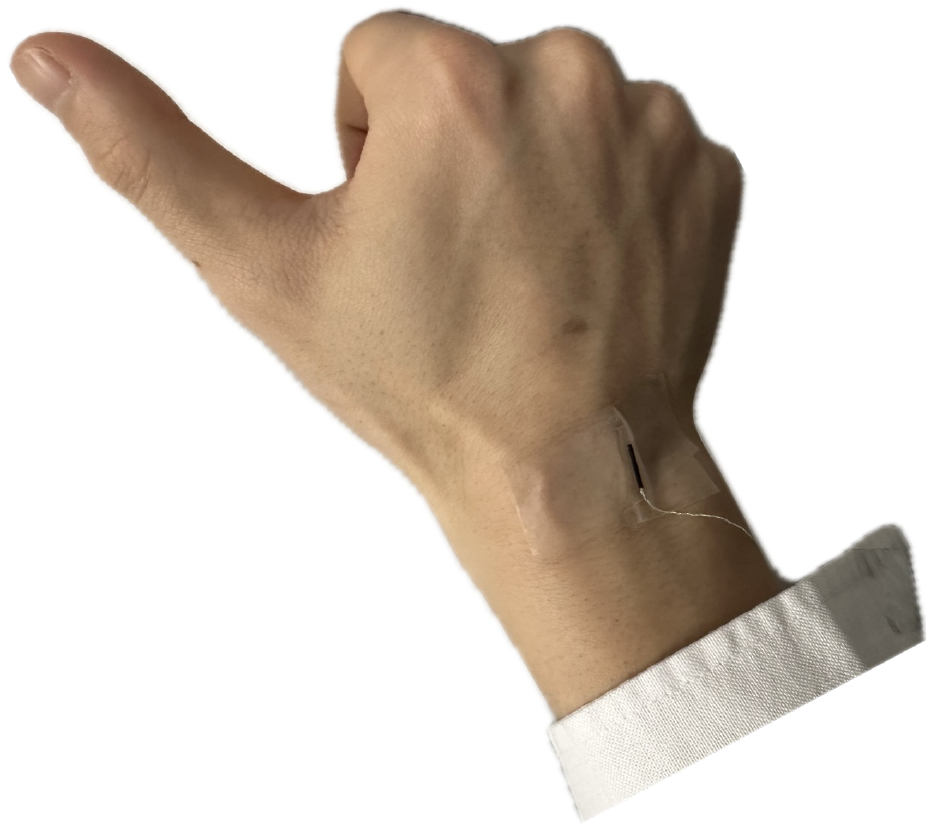}
\begin{tikzpicture}
\coordinate (ref) at (0,0);
\draw[<-, ultra thick, green] (ref)+(0.8,0.7) -- (0.8,2);
\filldraw[blue] (0.8,0.5) circle (2pt) node[anchor=west]{};
\end{tikzpicture}
\end{overpic}

\begin{tikzpicture}
    \begin{groupplot}[
        group style={
            group size=3 by 1,
            horizontal sep=0.8cm,
        },
        width=0.39\linewidth,
        height=0.4\linewidth,
        xmin=0.025, xmax=0.75,
        ymin=0.4, ymax=0.9,
        xtick={0.025,0.75},
        ytick={0.4,0.9},
        xticklabel style={
            anchor=near xticklabel,
            /pgf/number format/precision=3,
                /pgf/number format/fixed}, 
    y label style={at={(axis description cs:0.3,.5)},anchor=south},
           xlabel style={at={(axis description cs:0.5,0.1)},anchor=north}, 
    ]
    
\nextgroupplot[
    ylabel={\footnotesize Accuracy ($\%$)},
    title={Button $1$ and $2$},
    title style={fill=red!20, text=red, draw=black, very thick, rounded corners},
    axis lines=left,
]    \addplot[blue] plot [error bars/.cd, y dir=both, y explicit] table [y] {h1.dat};\label{plots:plot1}
    \addplot[red] plot [error bars/.cd, y dir=both, y explicit] table [y] {h2.dat};\label{plots:plot2}

\nextgroupplot[
    title={Button $2$ and $3$},
    title style={fill=red!20, text=red, draw=black, very thick, rounded corners},
    axis lines=left,ytick=\empty
]    \addplot[blue] plot [error bars/.cd, y dir=both, y explicit] table [y] {hh1.dat};\label{plots:plot1}
    \addplot[red] plot [error bars/.cd, y dir=both, y explicit] table [y] {hh2.dat};\label{plots:plot2}

\nextgroupplot[
    title={Button $3$ and $4$},
    title style={fill=red!20, text=red, draw=black, very thick, rounded corners},
    axis lines=left,ytick=\empty
]    \addplot[blue] plot [error bars/.cd, y dir=both, y explicit] table [y] {hhh1.dat};\label{plots:plot1}
    \addplot[red] plot [error bars/.cd, y dir=both, y explicit] table [y] {hhh2.dat};\label{plots:plot2}
    \end{groupplot}
    

\end{tikzpicture}
\usepgfplotslibrary{groupplots}
\begin{tikzpicture}
    \begin{groupplot}[
        group style={
            group size=3 by 1,
            horizontal sep=0.8cm,
        },
        width=0.39\linewidth,
        height=0.4\linewidth,
        xmin=0.025, xmax=0.75,
        ymin=-20, ymax=210,
        xtick={0.025,0.75},
        ytick={-20,210},
        xticklabel style={
            anchor=near xticklabel,
            /pgf/number format/precision=3,
            /pgf/number format/fixed},
    y label style={at={(axis description cs:0.3,.6)},anchor=south},
           xlabel style={at={(axis description cs:0.5,0)},anchor=north}, 
    ]
    
    \nextgroupplot[xlabel={Time (s)}, ylabel={$D$},axis lines = left]
    \addplot[blue] plot [error bars/.cd, y dir=both, y explicit] table [y] {d1.dat};
    \addplot[red] plot [error bars/.cd, y dir=both, y explicit] table [y] {d2.dat};
\addplot [name path=upper,draw=none] table[x=x,y expr=\thisrow{y}+\thisrow{error}/2.5] {d1.dat};
\addplot [name path=lower,draw=none] table[x=x,y expr=\thisrow{y}-\thisrow{error}/2.5] {d1.dat};
\addplot [fill=blue!10] fill between[of=upper and lower];
\addplot [name path=upper,draw=none] table[x=x,y expr=\thisrow{y}+\thisrow{error}/2.5] {d2.dat};
\addplot [name path=lower,draw=none] table[x=x,y expr=\thisrow{y}-\thisrow{error}/2.5] {d2.dat};
\addplot [fill=red!10] fill between[of=upper and lower];

    \nextgroupplot[xlabel={Time (s)},axis lines = left,ytick=\empty]
    \addplot[blue] plot [error bars/.cd, y dir=both, y explicit] table [y] {dd1.dat};
    \addplot[red] plot [error bars/.cd, y dir=both, y explicit] table [y] {dd2.dat};
\addplot [name path=upper,draw=none] table[x=x,y expr=\thisrow{y}+\thisrow{error}/2.5] {dd1.dat};
\addplot [name path=lower,draw=none] table[x=x,y expr=\thisrow{y}-\thisrow{error}/2.5] {dd1.dat};
\addplot [fill=blue!10] fill between[of=upper and lower];
\addplot [name path=upper,draw=none] table[x=x,y expr=\thisrow{y}+\thisrow{error}/2.5] {dd2.dat};
\addplot [name path=lower,draw=none] table[x=x,y expr=\thisrow{y}-\thisrow{error}/2.5] {dd2.dat};
\addplot [fill=red!10] fill between[of=upper and lower];

    \nextgroupplot[xlabel={Time (s)},axis lines = left,ytick=\empty]
    \addplot[blue] plot [error bars/.cd, y dir=both, y explicit] table [y] {ddd1.dat};
    \addplot[red] plot [error bars/.cd, y dir=both, y explicit] table [y] {ddd2.dat};
\addplot [name path=upper,draw=none] table[x=x,y expr=\thisrow{y}+\thisrow{error}/2.5] {ddd1.dat};
\addplot [name path=lower,draw=none] table[x=x,y expr=\thisrow{y}-\thisrow{error}/2.5] {ddd1.dat};
\addplot [fill=blue!10] fill between[of=upper and lower];
\addplot [name path=upper,draw=none] table[x=x,y expr=\thisrow{y}+\thisrow{error}/2.5] {ddd2.dat};
\addplot [name path=lower,draw=none] table[x=x,y expr=\thisrow{y}-\thisrow{error}/2.5] {ddd2.dat};
\addplot [fill=red!10] fill between[of=upper and lower];
    \end{groupplot}
    
   \path (group c1r1.south) -- node[midway, below, yshift=-10mm,xshift=12mm] {
        \ref{plots:plot1} $\dr$
        \ref{plots:plot2} $\df$
    } (group c2r1.south);

\node at (-0.8,-0.8){\ref{f:human_reaching-b}};
\node at (2.5,-0.8){\ref{f:human_reaching-c}};
\node at (5.2,-0.8){\ref{f:human_reaching-d}};
\end{tikzpicture}
\begin{tikzpicture}
    \begin{groupplot}[
        group style={
            group size=2 by 1,
            horizontal sep=0.8cm,
        },
        width=0.55\linewidth,
        height=0.5\linewidth,
        xmin=0.025, xmax=0.75,
        ymin=-36, ymax=-22,
        xtick={0.025,0.75},
        ytick={-36,-22},
        xticklabel style={
            anchor=near xticklabel,
            /pgf/number format/precision=3,
            /pgf/number format/fixed
        },      y label style={at={(axis description cs:0.3,.5)},anchor=south},
  every axis x label/.style={at={(current axis.south)},below=0.5mm}
    ]
    
    \nextgroupplot[xlabel={Time (s)}, ylabel={\footnotesize Vertical position(\si{\cm})}]
    \addplot[blue, line width=1.5pt] plot [error bars/.cd, y dir=both, y explicit] table [y] {ooo1.dat};\label{plots:plot5}
    \addplot[red, line width=1.5pt] plot [error bars/.cd, y dir=both, y explicit] table [y] {ooo2.dat};\label{plots:plot6}
    \addplot[blue!50] plot [error bars/.cd, y dir=both, y explicit] table [y] {ooo3.dat};\label{plots:plot7}
    \addplot[red!50] plot [error bars/.cd, y dir=both, y explicit] table [y] {ooo4.dat};\label{plots:plot8}
    
    \nextgroupplot[xlabel={Time (s)},ytick=\empty]
        \addplot[
        blue,          
        mark=none,     
        smooth ,
        line width=1.5pt
    ] table [y=class] {
        x   class
0.0250000000000000	-29.3854000000000
0.0500000000000000	-29.4302000000000
0.0750000000000000	-29.3103000000000
0.100000000000000	-29.4663000000000
0.125000000000000	-29.7524000000000
0.150000000000000	-29.6739000000000
0.175000000000000	-29.4942000000000
0.200000000000000	-29.5460000000000
0.225000000000000	-29.4233000000000
0.250000000000000	-29.4414000000000
0.275000000000000	-29.3122000000000
0.300000000000000	-29.0451000000000
	
    };

      \addplot[
        red,          
        mark=none,     
        smooth,
      line width=1.5pt
    ] table [y=class] {
        x   class
0.0250000000000000	-28.5857000000000
0.0500000000000000	-28.2698000000000
0.0750000000000000	-27.9118000000000
0.100000000000000	-27.8582000000000
0.125000000000000	-27.6815000000000
0.150000000000000	-27.4433000000000
0.175000000000000	-27.3536000000000
0.200000000000000	-27.0922000000000
0.225000000000000	-26.6747000000000
0.250000000000000	-26.4913000000000
0.275000000000000	-26.2096000000000
0.300000000000000	-25.8240000000000
    };

    \addplot[blue,dashed, line width=1.5pt] plot [error bars/.cd, y dir=both, y explicit] table [y] {rrr1b1.dat};\label{plots:plot9}
    \addplot[red, dashed,line width=1.5pt] plot [error bars/.cd, y dir=both, y explicit] table [y] {rrr1r1.dat};\label{plots:plot10}

\addplot [name path=upper,draw=none] table[x=x,y expr=\thisrow{y}+3*\thisrow{err}] {rrr1b1.dat};
\addplot [name path=lower,draw=none] table[x=x,y expr=\thisrow{y}-3*\thisrow{err}] {rrr1b1.dat};
\addplot [fill=blue!10] fill between[of=upper and lower];

\addplot [name path=upper,draw=none] table[x=x,y expr=\thisrow{y}+3*\thisrow{err}] {rrr1r1.dat};
\addplot [name path=lower,draw=none] table[x=x,y expr=\thisrow{y}-3*\thisrow{err}] {rrr1r1.dat};
\addplot [fill=red!10] fill between[of=upper and lower];

    \end{groupplot}
    
  \path (group c1r1.south west) -- (group c2r1.south east)
    node[midway, below, yshift=-5mm,xshift=-5mm] { 
        \setlength{\tabcolsep}{1.5pt} 
      \begin{tabular}{lll}
        \ref{plots:plot5} \footnotesize$\dr$ (Button 3) &  \footnotesize \ref{plots:plot6} $\dr$(Button 3)  & \ref{plots:plot9}  \footnotesize Future$\dr$ (Button 3) \\
        \ref{plots:plot7}  \footnotesize$\df$ (Button 2) &  \ref{plots:plot8}  \footnotesize $\df$(Button 2) &  \ref{plots:plot10} \footnotesize Future$\dr$ (Button 2)\\
      \end{tabular}
    };

    \draw[->, line width=1pt,black] (5.3,0.7) -- (5.3,1.1); 
\node at (5.6,0.5){$\dr:66\%$};
\node [text=red]at (5.6,0.2){$\df:82\%$};
      \node at (-0.8,-0.6){\ref{f:human_reaching-e}};
\node at (4,-0.6){\ref{f:human_reaching-f}};
\end{tikzpicture}
\caption{\cl{\item\label{f:human_reaching-a}Experimental setup. The participant presses a start button and then reaches for one of the target buttons. The sensors are attached to the wrist and the tip of the sleeve. The motion prediction accuracy and the cross-fitness distance between 
\item\label{f:human_reaching-b} button $1$ and $2$, \item\label{f:human_reaching-c} button $2$ and $3$ and \item\label{f:human_reaching-d} button $3$ and $4$ given various durations of the past movement. Reported are the mean value $\pm$ $\text{s.d}/5$.  \item\label{f:human_reaching-e} The vertical position of the sensor's movement when the human arm is reaching button $2$ and button $3$.  \item\label{f:human_reaching-f} The future movement of the human body (dashed line and shaded area) is predicted based on their past movement (solid line) from the initial time step up to \SI{0.6}{\second}. The future trajectory shown is the mean value (dashed line) $\pm$ $\text{s.d}\times5$ (shaded area). The motion prediction accuracies using the past movement until $\SI{1}{\second}$ recorded by sensor $\dr_{1}$ and $\df_{4}$ are shown.}}
\label{f:human_reaching}
    \begin{tikzpicture}[remember picture,overlay]
    \coordinate (bottomLeft) at ([xshift=-72mm,yshift=-72mm]current page.north east); 
    \coordinate (topRight) at ([xshift=-105mm,yshift=-135mm]current page.north east); 
    \draw[line width=1pt, rounded corners=1pt, draw=black] (bottomLeft) rectangle (topRight);
\end{tikzpicture}
    \begin{tikzpicture}[remember picture,overlay]
    \coordinate (bottomLeft) at ([xshift=-72mm,yshift=-72mm]current page.north east); 
    \coordinate (topRight) at ([xshift=-46mm,yshift=-135mm]current page.north east); 
    \draw[line width=1pt, rounded corners=1pt, draw=black] (bottomLeft) rectangle (topRight);
\end{tikzpicture}
   \begin{tikzpicture}[remember picture,overlay]
    \coordinate (bottomLeft) at ([xshift=-20mm,yshift=-72mm]current page.north east); 
    \coordinate (topRight) at ([xshift=-46mm,yshift=-135mm]current page.north east); 
    \draw[line width=1pt, rounded corners=1pt, draw=black] (bottomLeft) rectangle (topRight);
\end{tikzpicture}
\end{figure}

The experimental setup includes five buttons and five LED lights. The white button marks the hand's starting position, while the other four buttons represent different target positions. The distance between each target button and the start button is $\SI{45}{\cm}$, ensuring uniform travel durations to reach each target. The angle between two adjacent target buttons is $5$ degrees, so all target buttons are positioned on an arc centred on the start button. To deactivate a target LED light, the participant must press the button of the corresponding colour. \fref{human_reaching}\ref{f:human_reaching-a} shows the experimental setup.

The participant was provided with an information sheet and signed a consent form before visiting the laboratory. Before data collection, the participant's sitting position is standardised. The distance between the centre of the shoulder joint and the reference point (see \fref{human_reaching}\ref{f:human_reaching-a}) is $70\%$ of the participant's arm length. The sleeve is adjusted to ensure the cuff is aligned flush with the wrist using tape. All data collection is conducted exclusively in an audio-visual isolated room. To familiarise the participant with the movement, a brief video is presented to demonstrate the specific movements to be performed.

The participant is asked to remove or roll up the sleeve of any clothing on their arms and to wear an instrumented shirt ($100\%$ woven cotton) with one sensor (denoted as $\df$) attached to the cuff of the right-side sleeve (with a maximum possible displacement from the wrist of $\pm \SI{10}{\cm}$) and another sensor attached near the unciform bone (denoted as $\dr$). The participant is asked to clench the dominant hand into a fist and use the thumb to fully press the start button, which turns off the corresponding LED light. Subsequently, the subject reaches directly for and presses the button corresponding to the randomly illuminated LED within $2.5\si{\second}$. The participant's hand then returns to the starting position, awaiting the re-illumination of the LED light. This procedure is repeated $40$ times, with each target LED illuminating $10$ times per block. The experiment involved a total of $5$ blocks. The release times of all buttons are recorded.

Following this procedure, the healthy male participant is invited to this experiment. As a result, an equal number of samples ($50$ for each target button) are recorded. The start time index of each trajectory is identified when the change in the horizontal dimension of wrist-attached sensor movement exceeds $\SI{10}{mm}$. The length of each hand-reaching trajectory is determined by the minimal time interval between the release of the start button and the target buttons (\ie $\Tk = \SI{0.75}{\second}$). Each trajectory occurring within this designated time interval, initiating from each start time index, is retained for further analysis ($\Tv = \fk\times\Tk = 0.75\times40 = 30$ data points). Human motion prediction and trajectory forecasting are solved according to the methods outlined in \sref{hmm}.

\subsubsection{Results}
\fref{human_reaching} \ref{f:human_reaching-a} and \ref{f:human_reaching-b} show the motion prediction accuracies and the cross-fitness distance between buttons $1$ and $2$, buttons $2$ and $3$ and buttons $3$ and $4$, respectively. The motion prediction accuracy of the sleeve-attached sensor is higher than the wrist-attached sensor, and the former requires a shorter duration to attain high accuracy. During the beginning of the human reaching movement, the accuracy of the sleeve-attached sensor does not exhibit a significant increase. This may be attributed to the relatively low arm moving speed at the beginning of the movement. The cross-fitness distance of the sleeve-attached sensor is higher than the wrist-attached sensor. This shows the sleeve movement has additional information to simplify the prediction task. \fref{human_reaching}\ref{f:human_reaching-f} shows the future human movement (dashed line) forecasted based on the past sleeve movement given data from time step one ($\tv=1$, $\tk=0$) to $\tk=\SI{0.3}{\second}$.

\section{Discussion}\label{s:discussion}%
This work compares the motion prediction performance between using body-attached and clothing-attached sensors. Motion prediction is an important step before forecasting the future human movement using the body-attached sensor. Counter-intuitively, \il{ \item the performance of motion prediction improves using clothing movement and \item clothing-attached sensors require a shorter duration of the past trajectory to achieve high accuracy.\item This phenomenon is explained by computing the statistical distance (\ie cross-fitness distance). Motion artefacts can increase statistical distance which contains more discrimination information and makes it easier to distinguish two categories of movements.} 

This finding suggests that the sensor should be attached to loose clothing that can more effectively address the motion prediction problem using the body-attached sensor. Motion prediction in this study is primarily based on measurements of the orientation of the fabric since this is the primary sensing modality available from accelerometers/\gls{imu}s used in most wearable motion capture technologies, and therefore the findings may have widespread applications.

In turn, this finding could have implications for many applications in robotics and automation. Such as prostheses \cite{pitou2018embroidered}, and human-robot collaboration (\eg assist the worker in the industry \cite{li2023self}, the mobile robot could assist the staff in the retail environment \cite{chen2022human}). Future work may explore more complex human movements, such as multi-joint upper limb movement or gait.


%
{\scriptsize\printbibliography}%
\InputIfFileExists{sandpit.tex}{}{}%
\end{document}